\documentclass[10pt,twocolumn,letterpaper]{article}

\usepackage[pagenumbers]{wacv} 

\usepackage{subcaption}

\usepackage{tikz}
\usepackage{graphicx}
\usepackage{caption}
\usepackage{bm}
\usetikzlibrary{positioning}
\usepackage{algorithm, algpseudocode}

\definecolor{wacvblue}{rgb}{0.21,0.49,0.74}
\usepackage[pagebackref,breaklinks,colorlinks,allcolors=wacvblue]{hyperref}

\def\wacvPaperID{3541} 
\def\confName{WACV}
\def\confYear{2027}

\title{Annotator-Free Object Detection via Foundation Model Collaboration}

\author{Yasiru Ranasinghe, and Vishal M. Patel\\
Johns Hopkins University\\
{\tt\small \{dranasi1, vpatel36\}@jhu.edu}
}

\begin{document}
\twocolumn[{%
  \maketitle
  \vspace{-0.75em} 
  \begin{center}
    \includegraphics[width=\linewidth]{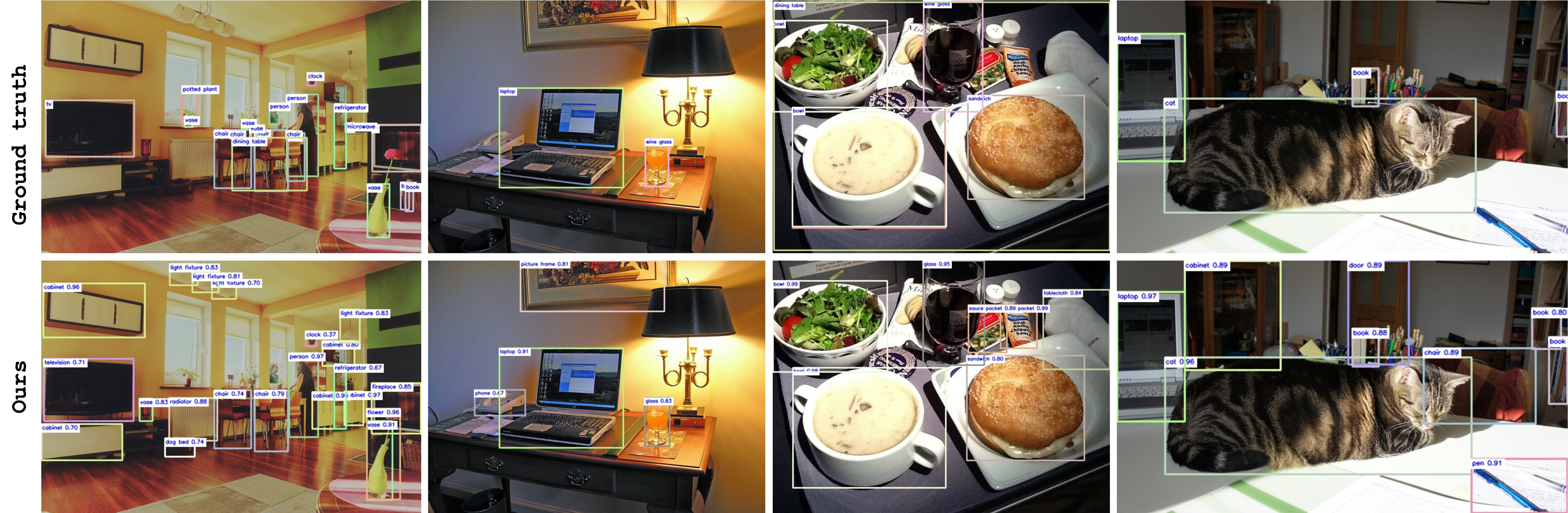} 
    \captionof{figure}{Qualitative results on COCO 2017 val. Top row: ground-truth annotations. Bottom row: predictions from our label-free detector, obtained by coordinating foundation models without any detection-specific supervision. The proposed method (1) detects most object classes present in the ground truth, even when expressed with synonymous labels (e.g., television for tv); (2) recovers objects that are missing annotations; and (3) proposes novel classes with no close counterpart in the dataset taxonomy (e.g., dog bed, radiator, pen, cabinet). For evaluation, a detection is considered correct if it has IoU $\geq$ 0.5 with a ground-truth box and a matching class (accounting for synonym mapping).
    }
    \label{fig:teaser}
  \end{center}
  \vspace{0.5em} 
}]
\begin{abstract}
Object detection research remains constrained by high annotation costs and the closed-set assumptions of supervised models, limiting scalability to novel categories. We present a target-dataset annotation-free object detection framework that eliminates the need for human labeling on the target dataset by leveraging the complementary strengths of pretrained foundation models, including segmentation and vision-language systems, combined with a self-refining training loop. To further reduce reliance on expensive LVLM queries, we introduce a landmark-based clustering mechanism that amortizes labeling across image clusters, improving label consistency while lowering computational cost. Our method autonomously generates segmentation proposals, assigns semantic labels through multimodal reasoning, and iteratively improves both localization and classification using a lightweight open-set detector. This modular pipeline enables adaptation to completely unseen object classes and domains without any human-provided target-dataset annotations, while maintaining compatibility with existing detection architectures. Evaluations on COCO, LVIS, ODinW-35, and Roboflow-100 demonstrate substantial performance gains over state-of-the-art open-vocabulary and long-tail detection methods, achieving 60.8 AP on COCO and 47.4 AP on LVIS under zero-shot settings. Beyond strong benchmark results, our approach provides a scalable foundation for real-world detection systems in unknown environments, where annotation is infeasible.
\end{abstract} 
\section{Introduction and Related Work}

\begin{figure}[!t]
    \centering
    \includegraphics[width=1\linewidth]{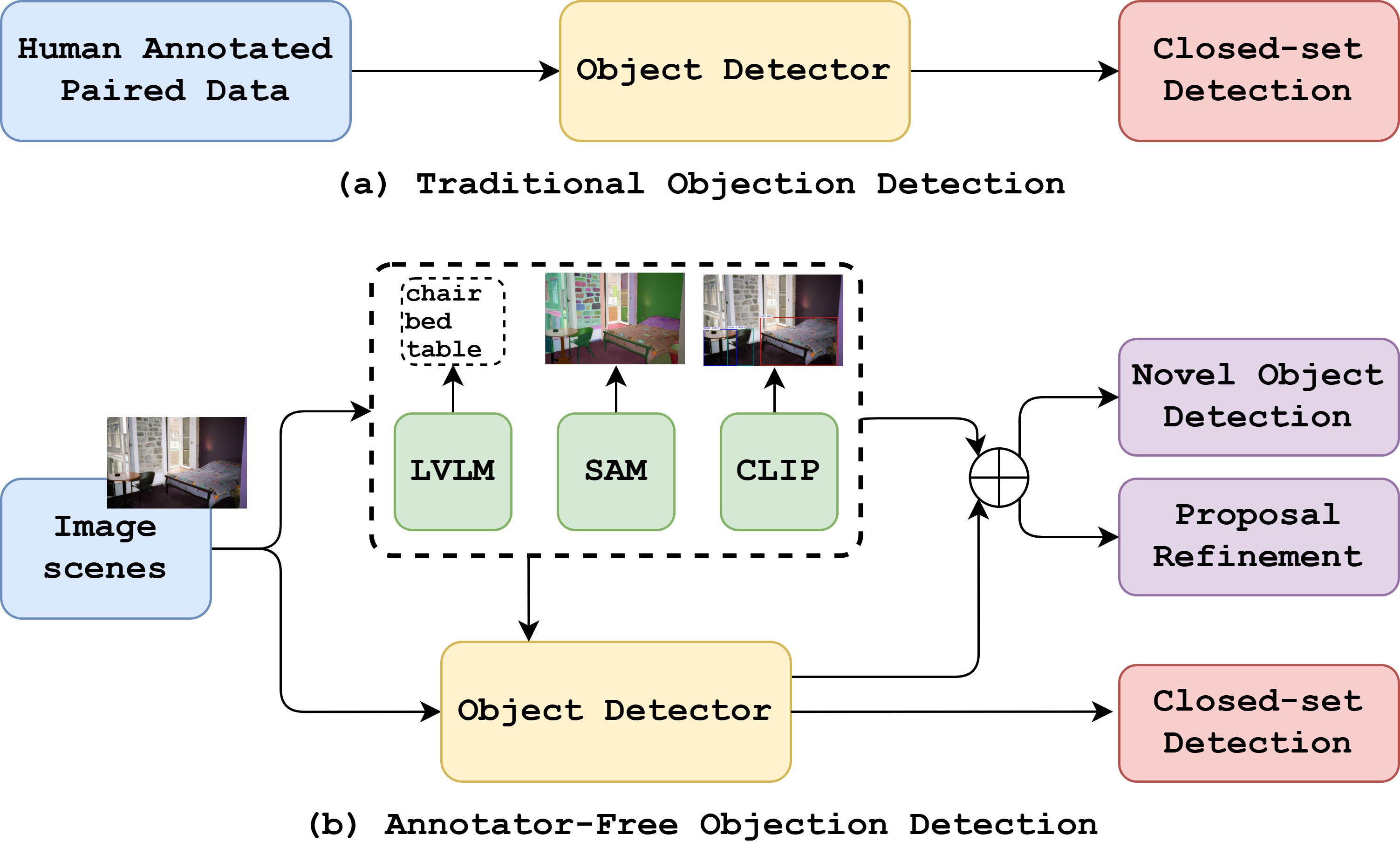}
    \caption{Traditional object detection requires annotated paired data and remains closed-set. Our annotator-free pipeline relies only on image scenes, generating proposals via foundational models and refining them iteratively to expand to novel categories.}
    \label{fig: highlight figure}
\end{figure}

Object detection is a fundamental task in computer vision with applications in autonomous driving, industrial automation, and surveillance \cite{od-survey, driving-autono, wang2023multi}. State-of-the-art detectors such as YOLO \cite{yolo} and DETR \cite{detr} achieve strong performance but rely heavily on annotated datasets, requiring precise bounding boxes and labels for every instance. This annotation process can take 5–30 minutes per image for complex datasets like COCO \cite{coco}, making large-scale training prohibitively expensive. Furthermore, detectors trained on such datasets are constrained to predefined categories, limiting their capacity to recognize novel or long-tail objects \cite{li2020overcoming, vaze2022generalized}. This challenge is illustrated in \cref{fig: highlight figure}.

Recent directions attempt to reduce annotation dependency. \emph{Novel object detection} explores recognizing unseen classes \cite{zhao2019object, rahman2018zero}. RNCDL \cite{fomenko2022learning} takes a self-supervised approach but cannot assign semantic labels without validation data. Generalized Class Discovery \cite{vaze2022generalized} and PromptCAL \cite{zhang2023promptcal} apply contrastive learning to categorize novel classes, yet they neglect localization \cite{rizve2022openldn}. Other works \cite{du2022vos} generate outliers to better separate known from unknown classes, but practical deployment remains limited.

\emph{Open-vocabulary object detection (OVD)} leverages vision-language alignment to expand recognition beyond closed-set constraints. Methods such as GLIP \cite{li2021grounded}, GroundingDINO \cite{groundingdino}, and YOLO World \cite{yoloworld} integrate natural language to detect unseen categories. However, these approaches still depend on closed-set detectors pre-trained on human-annotated data, keeping annotation costs high. While VLMs like CLIP \cite{clip} bridge vision and text, their effectiveness relies on supervised training of the  detector.

\emph{Unsupervised object detection and discovery} methods aim to localize and, in some cases, segment objects without any human annotation, typically by exploiting the emergent structure of self-supervised representations. CutLER \cite{cutler} discovers and segments multiple objects per image using self-supervised features combined with iterative self-training, without relying on any labels. FreeSOLO \cite{freesolo} produces coarse object masks in a fully unsupervised manner and refines them through self-training for instance segmentation. TokenCut \cite{tokencut} and MaskDistill \cite{maskdistill} exploit normalized-cut and self-distillation signals from self-supervised vision transformers to discover salient object regions without supervision. Complementary to these purely visual approaches, RegionCLIP \cite{regionclip}, OWLv2 \cite{owlv2}, and Grounded-SAM \cite{groundedsam} pair region-level vision-language pretraining or grounding modules with segmentation models such as SAM \cite{sam} to enable open-vocabulary localization. While these methods substantially reduce annotation requirements, they either lack the ability to assign open-ended semantic labels (CutLER, FreeSOLO, TokenCut, MaskDistill) or still depend on some form of pre-existing detection or grounding supervision, or curated region-text pairs (RegionCLIP, OWLv2, Grounded-SAM). Our approach differs by coordinating segmentation, vision-language grounding, and LVLM reasoning entirely on the target dataset at training/inference time, without requiring any target-dataset annotations or a fine-tuned detection backbone.

\begin{figure*}[!t]
    \centering
    \includegraphics[width=0.85\linewidth]{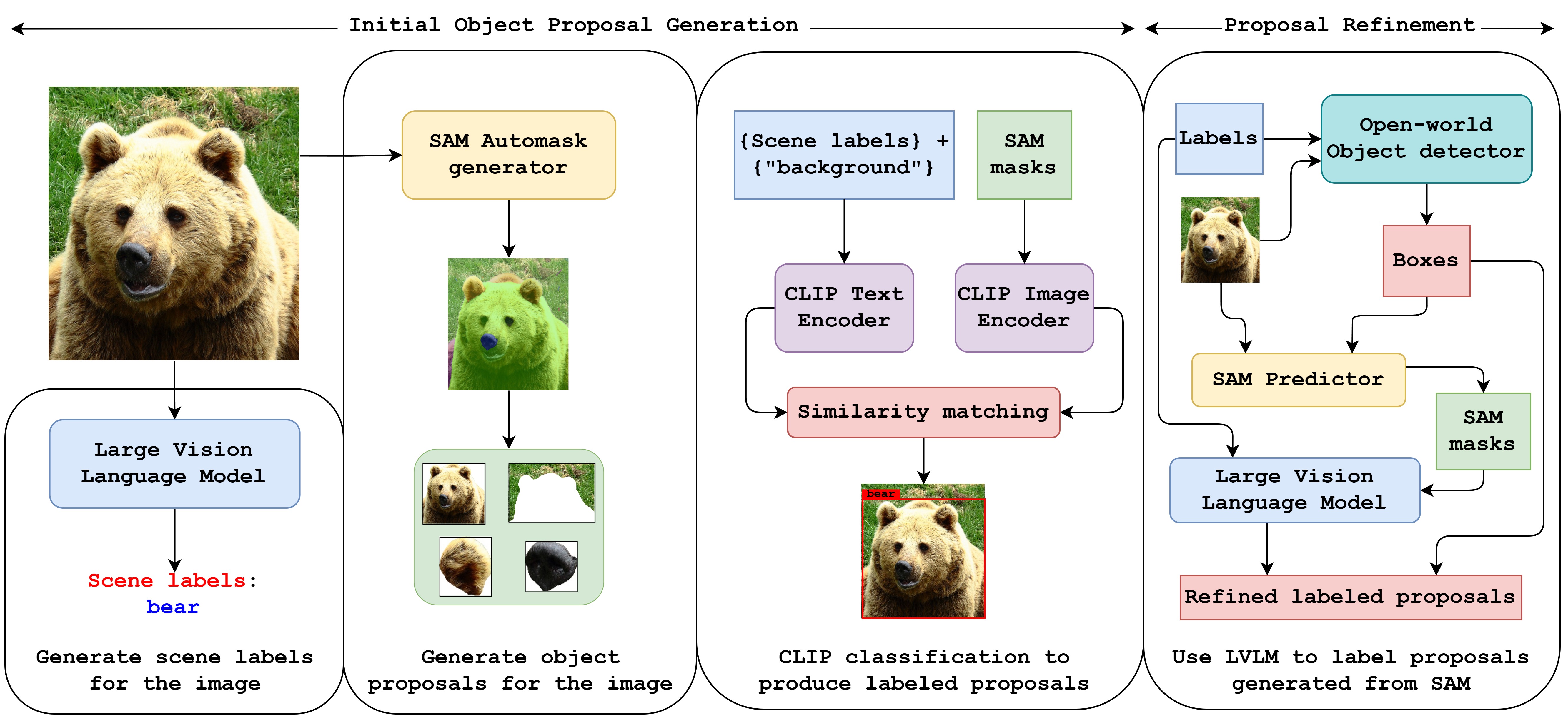}
    \caption{Proposed annotator-free pipeline. Stage 1: LVLM and SAM generate scene-level labels and object proposals. Stage 2: CLIP encoders assign semantic labels to proposals. Stage 3: an open-set detector trained on these proposals generates refined prompts for SAM, which are re-labeled by the LVLM.}
    \label{fig: pipeline}
\end{figure*}

In this work, we propose a \textbf{target-dataset annotation-free object detection} pipeline that eliminates the need for human annotation on the target dataset by adapting pretrained foundation models rather than removing human supervision altogether. Our approach integrates large vision-language models \cite{zhao2023survey} and foundational models such as SAM \cite{sam}, which are themselves trained on human-curated data, to autonomously generate object proposals and semantic labels for the target domain. As shown in \cref{fig: pipeline}, proposals are iteratively refined, enabling the detector to generalize to previously unseen object categories. Unlike OVD, which extends recognition but relies on annotated bases, our method directly adapts to novel environments without prior target-dataset labels. Benchmarks on LVIS and COCO validate that our pipeline reduces annotation costs and also has competitive performance in detecting novel categories.

\textbf{Our contributions are:}
\begin{enumerate}
    \item A novel target-dataset annotation-free detection pipeline integrating foundational and vision-language models, removing the need for human annotation on the target dataset.
    \item A self-refining mechanism that progressively enhances detection without labeled data.
    \item A modular design leveraging pre-trained models, enabling generalization to unseen categories.
    \item Extensive experiments showing superior results compared to state-of-the-art on challenging datasets.
\end{enumerate}

\section{Annotator-free object detection}
\label{sec: afod}
Annotator-free object detection removes the reliance on human-annotated ground truths, enabling detectors to train without prior category knowledge. In this setting, models encounter both known and unknown objects during inference. The goal is to eliminate annotation costs while improving detection of novel and existing classes. Unlike closed-set detectors that yield zero AP on unseen categories, our self-annotator mechanism, built on YOLO with support from LLMs, CLIP, and SAM, achieves gains in both novel and known AP.

Our approach follows three stages: \textbf{Self-labeling}, where foundation and language models generate bounding boxes and labels without human input, amortized via SAM{+}CLIP landmark clustering (\cref{sec: functionality breakdown}); \textbf{Detection}, where an open-set detector trained on these labels recognizes both seen and unseen classes; and \textbf{Refinement}, where SAM, guided by the trained detector, improves proposal quality and localization, as illustrated in \cref{fig: pipeline}.

\subsection{Deconstructing Dataset Preparation}
\label{sec: functionality breakdown}

\noindent\textbf{Landmark-based clustering before LVLM queries.}
To avoid querying an LVLM on every SAM proposal, we amortize labeling across \emph{images}. For image $I_i$, we run SAM to obtain masks $\{m_{ik}\}$ and form object crops by masking out the background (black by default). Each crop is encoded with CLIP to produce unit-normalized landmark features $\mathbf{z}_{ik}\in\mathbb{R}^{768}$. We keep up to $K{=}32$ largest valid masks per image, yielding the set $Z_i=\{\mathbf{z}_{ik}\}_{k=1}^{K_i}$ ($K_i\le K$). 

\noindent\textit{Neighbor shortlist.} To avoid all-pairs, we compute exact cosine kNN on per-image means
$\bar{\mathbf{z}}_i \!=\! \frac{1}{K_i}\sum_k \mathbf{z}_{ik}$ (then $\ell_2$-normalize), keeping $K_{\text{short}}{=}75$ candidate neighbors.

\noindent\textit{Image-to-image distance.} Given two images $i,j$ with landmark sets $Z_i$ and $Z_j$, we measure dissimilarity via a symmetric Chamfer over cosine distance:
\begin{equation}
\small
\begin{aligned}
d(i,j) = \tfrac{1}{2}\!\Bigg(&
\frac{1}{K_i}\!\sum_{\mathbf{a}\in Z_i}\min_{\mathbf{b}\in Z_j}\big(1-\mathbf{a}^{\top}\mathbf{b}\big) \\
&+\;
\frac{1}{K_j}\!\sum_{\mathbf{b}\in Z_j}\min_{\mathbf{a}\in Z_i}\big(1-\mathbf{a}^{\top}\mathbf{b}\big)
\Bigg).
\end{aligned}
\end{equation}

\noindent\textit{Sparse landmark graph.} For an image $i$ we sort its neighbors by $d(i,j)$ and retain an edge $(i,j)$ iff all filters pass:
\begin{enumerate}[leftmargin=1.4em,itemsep=2pt,topsep=2pt]
\item \textbf{kNN:} $j$ lies among the top-$k$ nearest of $i$ by $d$;
\item \textbf{Local percentile:} $d(i,j)\le\tau_i$, where $\tau_i$ is the $p$-th percentile of $\{d(i,\ell)\}$ over the shortlist of $i$;
\item \textbf{Global cutoff:} $d(i,j)\le\tau_g$, the global percentile (e.g., median) over all shortlisted distances (optionally tightened via a shared-neighborhood constraint $|\mathrm{NN}(i)\cap\mathrm{NN}(j)|\ge s$, where $\mathrm{NN}(\cdot)$ are the Stage-B kNN lists).
\end{enumerate}
We symmetrize with a mutual requirement (keep $(i,j)$ only if both $i$ keeps $j$ and $j$ keeps $i$). Connected components of this sparse graph define image clusters.


\noindent\textit{Medoids and label propagation.} For each component $\mathcal{C}$ we select a medoid image by maximizing degree and, as a tie-break, minimizing the sum of in-cluster distances:
\[
i^\star \;=\; \arg\max_{i\in\mathcal{C}} \big(\deg(i),\; -\!\!\sum_{j\in\mathcal{C},\,j\neq i} d(i,j)\big).
\]
We query the LVLM \emph{only} on medoids to obtain scene/object labels and propagate those labels to all images within the same component, reducing LVLM usage from $O\!\left(\sum_i K_i\right)$ per-dataset proposals to $O(C)$ queries while improving intra-cluster label consistency. We optionally repeat this clustering-and-labeling procedure across iterations to improve coverage: after each round, we (i) prune noisy SAM masks and re-compute landmarks for uncertain images using the newly assigned labels, and (ii) optionally update the detector (YOLO) on the current pseudo-labels, before rebuilding the shortlist and graph.




\noindent\textbf{Producing Object Labels}. The initial step in annotator-free object detection is autonomously generating semantic labels for detected objects, removing reliance on manual annotation. We leverage large vision-language models (VLMs) such as GPT-4o and CogVLM, which act as general-purpose observers, autonomously recognizing and describing objects present in a scene without any prior exposure or explicit definition of existing classes. Given an input image region \(\mathbf{x}_r\), the visual encoder \(\mathcal{F}_{v}^{(\text{VLM})}\) and multimodal fusion module \(\mathcal{G}_{f}^{(\text{VLM})}\) directly produce text outputs:
\begin{equation*}
    \mathbf{T} = \mathcal{G}_{f}^{(\text{VLM})}\left(\mathcal{F}_{v}^{(\text{VLM})}(\mathbf{x}_r)\right),
\end{equation*}
where \(\mathbf{T}\) represents descriptive object labels or captions reflecting objects the VLM inherently recognizes, without predefined label sets -- enabling flexible labeling and novel category discovery.

\noindent\textbf{Producing Bounding Box Proposals}. The second step autonomously generates bounding box proposals to localize objects within images, using a segmentation-based proposal generation module (\(\mathcal{F}_{\text{seg}}\)) -- specifically SAM \cite{sam} -- that produces segmentation masks \(\mathcal{M} = \{m_i\}_{i=1}^{N}\), from which bounding boxes \(\mathcal{B} = \{b_i\}_{i=1}^{N}\) are derived. Because SAM is trained on vast and diverse data, it generalizes across classes and domains without fine-tuning, yielding precise, reliable bounding-box proposals even in completely unknown environments.

\noindent\textbf{Assigning Labels to Bounding Box Proposals}. After independently generating semantic labels and bounding box proposals, the next critical step involves matching these semantic labels to the bounding box proposals. Since segmentation masks lack inherent semantic labels, this matching step requires aligning visual embeddings of bounding boxes with textual embeddings of semantic labels. A suitable model for this matching step must operate within a unified embedding space for visual and textual data. Given visual embeddings  for each bounding box proposal  and textual embeddings  for candidates, the matching's given by:
\begin{equation*}
y_i = \underset{c \in \mathcal{C}}{\arg\max}\; \text{sim}(\bm{\phi}_{b}(b_i), \bm{\phi}_{l}(c)).
\end{equation*}
This matching strategy enables effective assignment of labels to proposals, balancing accuracy and inference.
Here
$b_i$ is the $i$-th proposal; 
$\mathcal{C}$ is the candidate label set including \texttt{background}; 
$\boldsymbol{\phi}_b(\cdot)\!\in\!\mathbb{R}^d$ is the image encoder  producing a unit-normalized embedding; 
$\boldsymbol{\phi}_\ell(\cdot)\!\in\!\mathbb{R}^d$ is the text encoder applied to a prompt producing a unit-normalized embedding; 
$\mathrm{sim}(\mathbf{u},\mathbf{v}) \!=\! \mathbf{u}^\top \mathbf{v}$ denotes cosine similarity for unit vectors; 
and $y_i$ is the assigned label for proposal $b_i$.

\subsection{Self-labeling for proposal generation}
The self-labeling stage involves generating initial bounding boxes as illustrated in \cref{fig: pipeline} and object labels for YOLO World. This stage combines several foundational models to complement each other’s abilities. SAM produces segmentation masks for a given scene, aiding in Region of Interest (ROI) or proposal generation. Large language models (LLMs) or large vision-language models (LVLMs) are capable of identifying objects present in the scene, helping to construct scene labels. CLIP provides strong zero-shot classification capabilities to ground object proposals from SAM with scene labels derived from LLMs.

SAM is a foundational segmentation model that can produce high-precision segmentation masks for objects in a scene. However, SAM requires input prompts to segment specific objects from the scene. Since there is no way to prompt SAM without a supervised object detector, SAM can be configured to produce all possible segmentation masks for the scene. This approach, however, results in unnecessary segmentation masks, including backgrounds, redundant masks, fine-grained parts, and noisy segmentations. To enable annotator-free object detection, these unnecessary segmentations should be removed to yield appropriate object proposals for detector training.

Large vision-language models are trained on vast datasets and have a large capacity for parameterized knowledge. This capacity enables LVLMs to perform zero-shot inference for objects present in a scene without prior exposure to specific training or testing distributions. However, due to their high capacity, LVLMs have significant inference times, which leads to high computational costs when filtering out unnecessary object proposals from SAM, making the pipeline impractical for real-time applications.

Since LVLMs have slow inference times for object labeling, we employ CLIP for pre-filtering. CLIP, a vision-language model, is smaller in size compared to both SAM and LVLMs. Given the input image proposals and text proposals, CLIP assigns labels from a predefined label set significantly faster than any LVLM. Additionally, due to CLIP's architecture, it produces confidence scores for each object proposal's labels, enabling benefits such as Non-Maximum Suppression for redundancy removal.

\subsection{Implementing the detector}

In the scene-level label generation, we begin by using the text prompt “Generate the list of general labels of the objects present in the scene.” However, this approach does not guarantee that the object labels generated for instances of the same object class will be consistent. For example, “television” and “tv” refer to the same object category but are linguistically different labels. This issue creates a run-off problem for object classes and leads to label redundancy. To mitigate this, we construct a dictionary to update the learned object categories. The dictionary is initialized without any prior knowledge of object classes, and it is updated as new categories are learned. Furthermore, for each object label produced by the LVLM, we search for similar labels or synonyms in the dictionary using the instruction “Replace the labels in the \{label list\} with similar labels present in the \{dictionary\}” to remove redundant labels. Because LVLM APIs (e.g., GPT-4o) are not guaranteed to produce deterministic outputs across repeated calls or successive passes over the dataset, we freeze the label dictionary \(\mathcal{D}\) once it converges (i.e., no new categories are added over a full pass through the dataset) and use this fixed dictionary for all subsequent proposal generation and evaluation. This prevents label-space drift caused by API non-determinism from affecting the reported results.

In the object-level label generation, we use text embeddings from the CLIP text encoder for scene-level labels and additionally append the class tag “background” to filter out unnecessary segmentations. To assign a label, we consider each segmentation instance: the segmented area is cropped using bounding boxes from SAM’s postprocessing. Then, the cropped image is multiplied by the binary segmentation mask to capture the region of interest (ROI) and exclude unnecessary parts from the cropped image. This step helps prevent background segmentation masks from being mislabeled as object labels, as the bounding box corresponding to the segmentation mask might include objects from other categories. Besides preventing background mislabeling, this also reduces cross-labeling of objects. 

Once the CLIP labels are obtained, objects labeled as “background” are removed from the object proposals. To further reduce redundancy among object proposals, we apply Non-Maximum Suppression using CLIP scores as the confidence values. After generating the final object proposals for the dataset, we train YOLO World to develop a supervised object detector capable of performing OWD.

\subsection{Proposal refinement}

Once the open-world object detector is trained, we use the scene-level labels to perform inference. However, this inference stage operates under a closed-set setting, meaning we only consider the categories included in the dictionary developed during the proposal generation phase. Next, we use the bounding boxes from YOLO World as box prompts for SAM to refine segmentations and eliminate unnecessary segmentations that were generated during the self-labeling process. The supervised detector significantly reduces the number of proposals generated for a scene while also improving the quality of SAM’s segmentation masks.

Due to the substantial reduction in object proposals, we employ an LVLM to generate refined labels for these proposals during the refinement stage. This step leverages the classification head of YOLO World, which uses the CLIP text encoder, making the classification head functionally equivalent to object-level label generation, as seen in the self-labeling stage. In this process, the LVLM acts as a validator for the labels generated by CLIP, yielding more accurate labels for object proposals. To ensure coherence between object-level labels and scene-level labels within the dictionary, we prompt the LVLM with the instruction, “Pick a label from the \{label list\} for the object,” which helps assign a consistent label. After refining the object proposals, we retrain the open-set object detector with the improved proposals, further enhancing detection performance. Algorithm~\ref{alg:self_labeling_training} summarizes the complete self-labeling and training process.

\begin{algorithm}[!t]
\caption{Self-Labeling and Training Process}
\label{alg:self_labeling_training}
\begin{algorithmic}[1]
\State \textbf{Input:} Image \( \mathbf{x} \)
\State \textbf{Initialize:} Label dictionary \( \mathcal{D} \)
\State Produce scene-level labels \( \mathcal{L}_{\text{scene}} \) for the image \( \mathbf{x} \) using LVLM
\State Realign labels \( \mathcal{L}_{\text{scene}} \) with the existing labels in the dictionary \( \mathcal{D} \) and update \( \mathcal{D} \) for new labels
\State Generate segmentation masks \( \mathcal{M} \) for the image \( \mathbf{x} \) using SAM
\State Classify the object proposals in \( \mathcal{M} \) using CLIP with scene-level labels \( \mathcal{L}_{\text{scene}} \)
\State Train YOLO World using the proposals generated for the dataset
\State Produce box prompts \( \mathbf{b}_{\text{prompt}} \) using the trained detector for SAM
\State Label the object proposals \( \mathcal{P} \) using LVLM to improve self-labeling quality
\State Retrain  detector with the improved object proposals \( \mathcal{P} \)
\end{algorithmic}
\end{algorithm}

\subsection{Novel object detection}
\paragraph{Process for Detecting Novel Object Classes.}
To detect novel object classes, we follow a process similar to the refinement stage. First, we generate scene-level labels for the scene and provide these labels to the open-set detector. The detector then produces regions of interest (ROIs), labels, and confidence scores for each detected object in the scene. From the scene-level labels, we separate the known and novel categories present in the scene. This allows us to focus on distinguishing high-confidence known categories from potential novel classes. For known categories, CLIP confidence scores tend to be higher for familiar class labels than for novel ones (reflecting CLIP's pretraining bias toward frequently-seen concepts, which yields tighter image-text alignment), aiding in differentiating known objects from unknown ones. For object proposals with low confidence scores, we extract the corresponding bounding boxes and follow the procedure from the refinement stage to label these as novel classes. This approach allows for accurate labeling of novel objects within the scene.

\section{Experiments and Results}
In \cref{sec: implementation details}, we outline the implementation settings for our study. We then present our main findings in \cref{sec: main results}, where we compare our approach to leading models, GDINO \cite{groundingdino} and T-Rex2 \cite{trex2}. Lastly, in \cref{sec: ablation}, we provide extensive ablation studies that highlight the practical advantages of our proposed approach.

\begin{figure}
    \centering
    \includegraphics[width=0.85\linewidth]{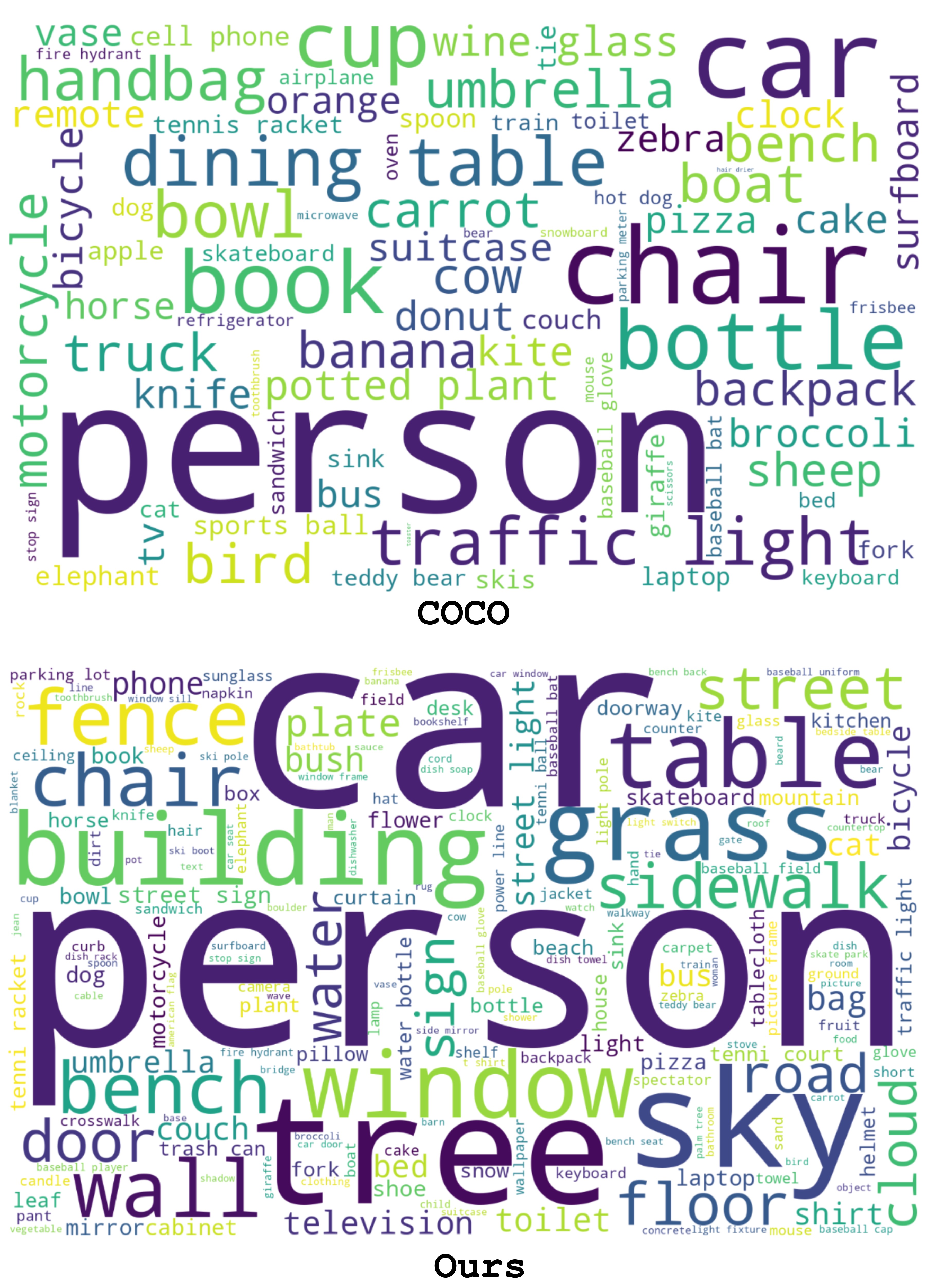}
    \caption{Word clouds of COCO ground-truth instances (top) and our pseudo-labeled annotations (bottom). The proposed annotation scheme preserves relative class frequencies (e.g., person, car, table) while introducing additional categories (e.g., fence, building, window), thereby enriching label diversity. This demonstrates that our method not only captures the underlying distribution but also scales to novel concepts beyond the original dataset.}
    \label{fig:wordcloud}
\end{figure}

\subsection{Implementation Details}
\label{sec: implementation details}

\textbf{Datasets:} We evaluate on five benchmarks: COCO \cite{coco}, LVIS \cite{gupta2019lvis}, ODinW-35 \cite{li2021grounded}, Roboflow-100 \cite{Ciaglia2022Roboflow1A}, and a COCO Open-Vocabulary Detection (OVD) split. COCO (2017 val) covers 80 classes, while the OVD split partitions 48 known and 17 novel classes. LVIS provides 1,203 categories for large-vocabulary evaluation. ODinW-35 and Roboflow-100 further test adaptability across varied domains, using their full validation splits without additional training data.

\begin{table}[!ht]
\centering
\small
\caption{Annotator-free object detection performance comparison on COCO val and LVIS minival. We consider the zero-shot evaluation of state-of-the-art long tail object detectors and open vocabulary object detectors for comparison.}
\label{tab: coco lvis results}
\begin{tabular}{@{}lccc@{}}
\toprule
\textbf{Model} & \textbf{Backbone} & \textbf{COCO val} & \textbf{LVIS} \\
\midrule
GLIP-L \cite{li2021grounded}          & Swin-L          & 49.8  & 26.9    \\
DINOv \cite{dino}           & Swin-L          & 46.2  & -         \\
GDINO-L \cite{groundingdino} & Swin-L          & 48.4  & 27.4    \\
T-Rex2 (Text) \cite{trex2}   & Swin-L          & 52.2  & 45.8   \\
T-Rex2 (Visual) \cite{trex2} & Swin-L        & 46.5  & 45.3   \\
DetCLIPv3 \cite{yao2024detclipv3} & Swin-T  & 47.2 & 47.0  \\
DetCLIPv3 \cite{yao2024detclipv3} & Swin-L  & 48.5 & 48.8  \\
Ours & Swin-T & 57.2 & 46.2 \\
Ours & YOLO       & 60.8  & 47.4 \\
\bottomrule
\end{tabular}
\end{table}

\textbf{LVLMs:} We employ two models: (i) \textbf{GPT-4o}, via API, for scene-level labeling, and (ii) \textbf{CogVLM}, open-source, for object-level labeling and for realigning scene labels with the label dictionary.

\textbf{YOLO World:} Built on MMYOLO/MMDetection, trained with AdamW (lr=0.002, wd=0.05) for 100 epochs on 8 NVIDIA A5000 GPUs (batch size 64). Following prior work, we apply color, affine, flip, and mosaic augmentations during pre-training.

\textbf{Evaluation protocol:} Unless otherwise noted, all quantitative results reported in \cref{tab: coco lvis results,tab: odinw and roboflow,tab: open-set object detection,tab: domain shift} use the standard AP@[.50:.95] metric under strict, exact category-name matching against each benchmark's ground-truth taxonomy; a prediction whose label does not exactly match a ground-truth category name is counted as a false positive even if it names a synonymous concept (e.g., ``television'' vs.\ ``tv''). This differs from the qualitative teaser in \cref{fig:teaser}, whose caption explicitly credits a detection as correct under IoU $\geq 0.5$ with a \emph{synonym-mapped} class label.

\subsection{Annotator-free object detection}
\label{sec: main results}

The performance on COCO val2017 and LVIS minival datasets is compared with state-of-the-art object detectors in \cref{tab: coco lvis results}. For a fair comparison, we evaluate all methods under a zero-shot setting, ensuring no prior access to the test set’s ground truth data. Our proposed method achieves a substantial improvement, with a 16.5$\%$ gain on COCO compared to long-tail detectors such as T-Rex2 and a 25.6$\%$ improvement over open-vocabulary detectors like GDINO. On LVIS, our method further surpasses long-tail detectors by 5$\%$ and open-vocabulary detectors by 72$\%$. The label distributions visualized in \cref{fig:wordcloud} further confirm that our pseudo-labeling scheme preserves COCO's class frequencies while introducing new categories. Qualitative examples are further illustrated in \cref{fig: qualitative comparisons}. To verify that these gains are not an artifact of the YOLO backbone, we additionally trained our pipeline using a Swin-T backbone \cite{liu2021swin}, achieving 57.2 AP on COCO and 46.2 AP on LVIS (\cref{tab: coco lvis results}), confirming that our pseudo-labeling pipeline provides substantial gains regardless of the detector architecture.

\begin{table}[!ht]
\centering
\small
\caption{Annotator-free object detection performance comparison on ODinW-35 val and Roboflow-100 val. We consider the zero-shot evaluation of state-of-the-art long tail object detectors and open vocabulary object detectors for comparison.}
\label{tab: odinw and roboflow}
\begin{tabular}{@{}lccc@{}}
\toprule
\textbf{Model} & \textbf{Backbone} & \textbf{ODinW} & \textbf{Roboflow} \\
\midrule
GLIP-L \cite{li2021grounded}          & Swin-L          & 23.4  & 8.6  \\
DINOv \cite{dino}           & Swin-L          & 15.7  & -    \\
GDINO-L \cite{groundingdino} & Swin-L          & 22.3  & 8.3  \\
T-Rex2 (Text)  \cite{trex2} & Swin-L          & 22.0  & 10.5 \\
T-Rex2 (Visual) \cite{trex2} & Swin-L        & 27.8  & 18.5 \\
Ours & Swin-T & 30.1 & 20.8 \\
Ours & YOLO       & 32.5  & 23.7 \\
\bottomrule
\end{tabular}
\end{table}

\begin{table}[!t]
\centering
\small
\caption{Comparison for novel object detection with other state-of-the-art methods. Performance is evaluated on a novel, known, and all classes on COCO OVD split following a generalized open-vocabulary detection setting.}

\label{tab: open-set object detection}
\begin{tabular}{@{}lcccc@{}}
\toprule
\textbf{Method} & \textbf{Backbone} & \textbf{Novel} & \textbf{Known} & \textbf{All} \\
\midrule
OVR-CNN \cite{ovr-cnn}         & RN50            & 22.8  & 46.0  & 39.9 \\
ViLD \cite{gu2022openvocabulary}                & ViT-B/32        & 27.6  & 59.5  & 51.3 \\
Detic \cite{detic}             & RN50            & 27.8  & 47.1  & 45.0 \\
OV-DETR   \cite{ov-detr}       & ViT-B/32        & 29.4  & 61.0  & 52.7 \\
BARON \cite{wu2023aligning}              & RN50            & 34.0  & 60.4  & 53.5 \\
CORA \cite{wu2023cora}                & RN50x4          & 41.7  & 44.5  & 43.8 \\
CoFoMo \cite{bharadwaj2023enhancing}   & ResNet101 & 50.3 & 49.8 & 49.9 \\
SAS-Det \cite{zhao2024taming}   & ResNet101 & 37.4 & 58.5 & 53.0 \\
DetCLIPv3 \cite{yao2024detclipv3} & Swin-T & 54.7 & 42.8 & 46.9 \\
Ours  & Swin-T & 56.3 & 62.7 & 58.3 \\
Ours                           & YOLO   & \textbf{55.9} & 64.6 & 59.1 \\
\bottomrule
\end{tabular}
\end{table}

The performance of our method on ODinW-35 val and Roboflow-100 val datasets is presented in \cref{tab: odinw and roboflow}. These datasets are designed to evaluate in-the-wild object detection performance. Our proposed method achieves a 17\% improvement over long-tail object detectors and a 45\% improvement over open-vocabulary object detectors on ODinW. On Roboflow, we observe even greater gains, with improvements of 28\% and 86\% for long-tail and open-vocabulary detection schemes, respectively. The ability to perform object detection in the wild without human annotations highlights the value of our pipeline, as it enables the development of automatic target recognition systems for unknown environments and niche object categories. The same Swin-T backbone described above achieves 30.1 AP on ODinW and 20.8 AP on Roboflow (\cref{tab: odinw and roboflow}), consistent with the backbone-parity trend observed on COCO and LVIS.

\begin{figure}[!ht]
    \centering
    \includegraphics[width=0.85\linewidth]{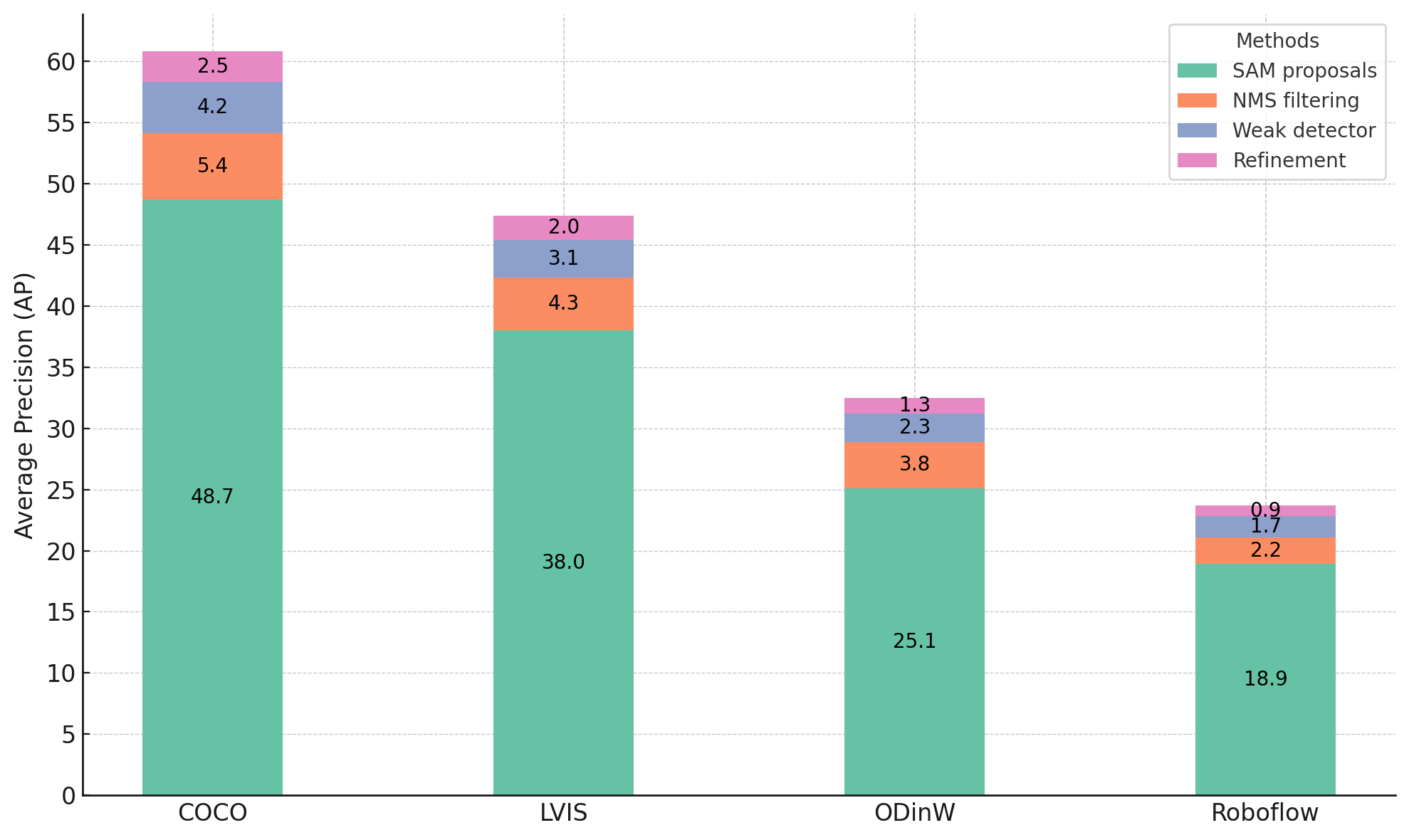}
    \caption{Ablation of performance improvement across pipeline stages (proposal generation, inference, refinement).}
    \label{fig: stage ablation}
\end{figure}

\subsection{Novel object detection}
\label{sec: novel object detection}

To evaluate the performance of our method on the COCO Open-Vocabulary Detection (OVD) split, we trained on 48 base COCO classes and tested it on 17 novel classes. To ensure a fair comparison with other methods, we trained the detector on the base classes using ground truth data for object detection, consistent with the approach taken by previous methods. Unlike existing OVD methods \cite{groundingdino}, our approach does not rely on any additional training datasets. Despite this, our method outperforms state-of-the-art approaches by a significant margin, as shown in \cref{tab: open-set object detection}. Specifically, we observe a 1.2-point improvement in AP\textsubscript{50} on novel classes (55.9 vs.\ 54.7) over the strongest competing method, DetCLIPv3 \cite{yao2024detclipv3}, underscoring the effectiveness of our approach relative to prior OVD methods. The Swin-T backbone variant introduced in \cref{sec: main results} shows a similar trend on this split, achieving 56.3/62.7/58.3 AP\textsubscript{50} on novel/known/all classes, respectively (\cref{tab: open-set object detection}).

Additionally, we analyzed the pipeline’s performance for novel classes using an object detector trained with our annotator-free detection approach. For the base classes, this detector achieved an AP\textsubscript{50} of 48.9, while for novel classes, we observed an AP\textsubscript{50} of 55.9. Unlike the results in \cref{tab: open-set object detection}, where the base-class detector is trained using \emph{ground-truth} annotations for fair comparison with prior OVD methods, the detector analyzed here is trained entirely through our annotator-free pipeline, without any ground-truth base-class labels; this is why its base-class AP\textsubscript{50} (48.9) differs from, and is not directly comparable to, the ground-truth-trained variant reported in \cref{tab: open-set object detection}. These results nonetheless demonstrate that with a sufficiently trained object detector, the proposed pipeline can effectively perform open-vocabulary detection without any ground-truth supervision at all.

\subsection{Ablation study}
\label{sec: ablation}

\begin{figure}[!ht]
    \centering
        
    
    \begin{subfigure}[b]{0.48\linewidth}
        \includegraphics[width=\linewidth]{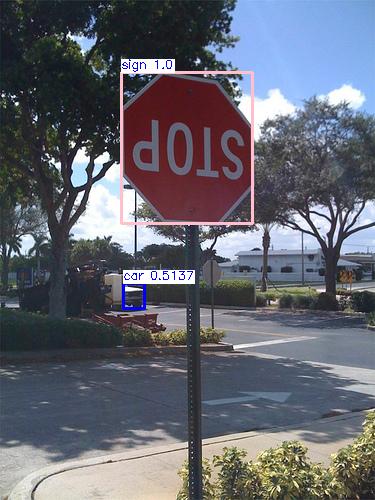}
    \end{subfigure}
    \begin{subfigure}[b]{0.48\linewidth}
        \includegraphics[width=\linewidth]{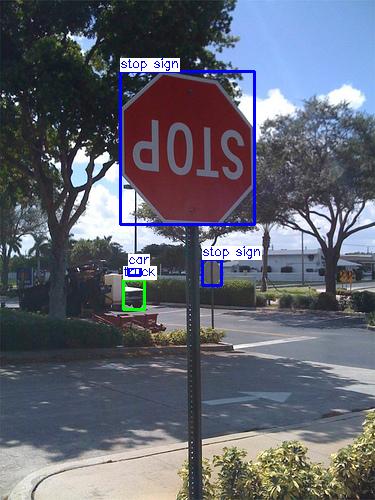}
    \end{subfigure}

    \begin{subfigure}[b]{0.48\linewidth}
        \includegraphics[width=\linewidth]{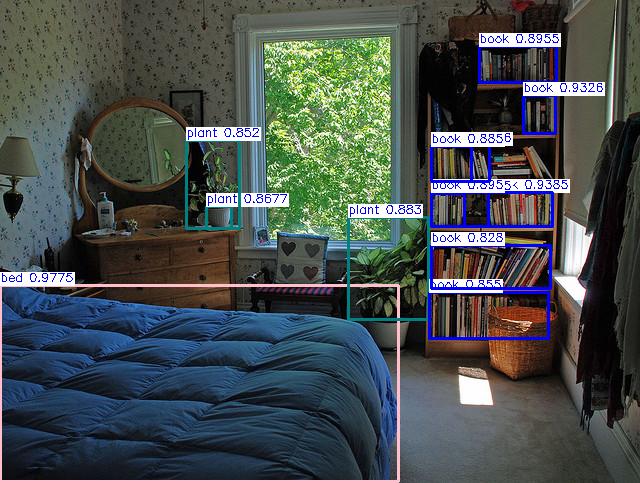}
    \end{subfigure}
    \begin{subfigure}[b]{0.48\linewidth}
        \includegraphics[width=\linewidth]{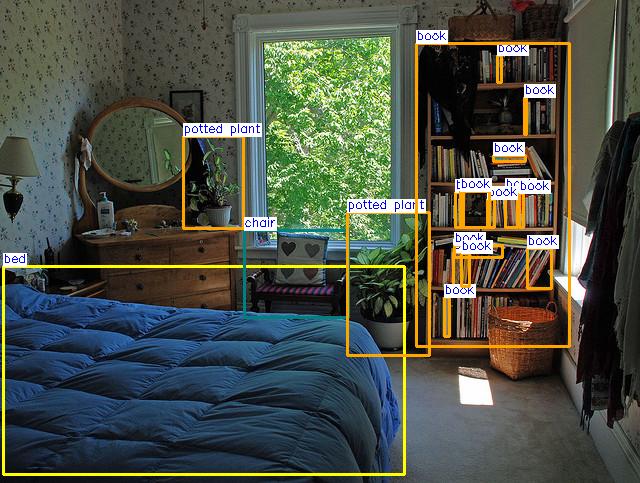}
    \end{subfigure}

    \caption{Qualitative results for the detection performance of the proposed annotator-free method. The predicted detection results are on the left column, and the ground truth results are on the right.}
    \label{fig: qualitative comparisons}
\end{figure}

\textbf{Performance gain at different stages}:
The object detection performance at different stages of the proposed method is shown in \cref{fig: stage ablation}. Results are provided for four datasets under an annotator-free detection scheme. For the ablation study, we evaluate the original object proposals generated by SAM, followed by proposals filtered using Non-Maximum Suppression (NMS) with CLIP scores. Next, we report detection performance from the trained object detector using the self-labeled dataset, and finally, we present the performance of the complete pipeline after refining the object proposal generation process.

As seen in \cref{fig: stage ablation}, the largest improvement is achieved at the original object proposal stage, demonstrating the effectiveness of foundation models in generating proposals for annotator-free object detection. Notably, this stage operates independently of any standard object detector and requires no hyperparameter tuning, making it a fully training-free process. Despite its simplicity, this stage yields performance on par with zero-shot evaluations and surpasses open-vocabulary detectors like GroundingDINO.

The performance gain observed in the NMS filtering stage is due to the removal of overlapping proposals generated in the prior stage. Subsequently, the detection performance after primary training on the object detector shows an incremental boost across all four datasets. This improvement is attributed to enhanced bounding box detection accuracy, indicating the generalization benefits of a trained detector over other training-free methods.

\begin{figure}[!ht]
    \centering
    \includegraphics[width=0.48\linewidth]{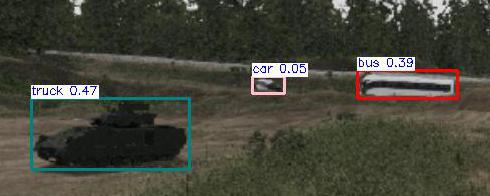}
    \includegraphics[width=0.48\linewidth]{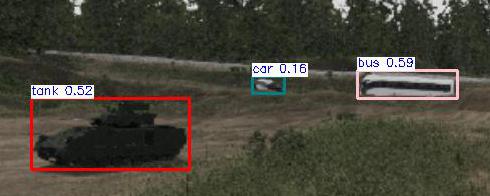}

    \vspace{0.15cm}

    \includegraphics[width=0.48\linewidth]{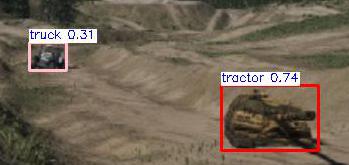}
    \includegraphics[width=0.48\linewidth]{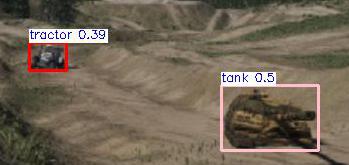}

\caption{Open-world detectors for novel object categories under domain shift. The detection performance for GroundingDINO (left column) are compared with the proposed novel object detection (right column).}
    \label{fig: domain shift}
\end{figure}

\noindent\textbf{Robustness under domain shift}: To test generalization beyond our four main benchmarks, we evaluate on a synthetic domain-shift dataset containing both known and novel categories unseen during training (Applied Intuition simulator; full setup and per-class results in the supplementary material, \cref{tab: domain shift}). As shown in \cref{fig: domain shift}, GroundingDINO can localize novel objects but often assigns incorrect classes, whereas our approach produces more accurate and consistent recognition. Our pipeline achieves 43.7/32.8/38.8 AP\textsubscript{50} on novel/known/overall classes, outperforming GLIP-L, T-Rex2, and GDINO-L.

\textbf{Computational cost}: Proposal generation is the main overhead, with LVLM queries costing roughly one second per scene. We avoid labeling all SAM segmentations directly, using CLIP as a pre-filter to cut labeling cost by $60\times$. Our landmark-based clustering (\cref{sec: functionality breakdown}) queries the LVLM only on cluster medoids and propagates labels across the cluster, dropping per-image labeling time from 15--20 minutes to $\sim$45 seconds ($\sim$20--27$\times$).

\subsection{Limitations}
\label{sec: limitations}

Despite strong results, our pipeline has notable limitations
(\cref{fig: qualitative comparisons,fig: domain shift}) such as: LVLM mislabeling of ambiguous or occluded objects propagates into the pseudo-labels used for training. Localization quality is bounded by SAM's segmentation, which degrades under low contrast, clutter, or fine-grained/overlapping parts. Labeling is also sensitive to LVLM prompt phrasing, only partially mitigated by the label dictionary (\cref{sec: functionality breakdown}). Finally, LVLM queries remain the dominant computational cost; our medoid-based clustering substantially reduces this (15--20 min to $\sim$ 45 s per image), but the label-quality trade-off remains uncharacterized.

\section{Conclusion}
We introduced a novel annotator-free object detection framework that leverages large language and foundational models to eliminate human labeling. By automatically generating proposals and refining them iteratively, our modular pipeline integrates seamlessly with existing open-set detectors, boosting performance on both known and novel classes. Beyond strong results in zero-shot and open-vocabulary settings, our approach demonstrates robustness under domain shifts, outperforming state-of-the-art methods such as GroundingDINO. These findings highlight the scalability and adaptability of our method, making it a practical solution for real-world detection in unstructured environments with minimal human intervention.

{
    \small
    \bibliographystyle{ieeenat_fullname}
    \bibliography{main}

@String(CVPR= {IEEE Conf. Comput. Vis. Pattern Recog.})

@String(ICCV= {Int. Conf. Comput. Vis.})

@String(CVPR  = {CVPR})

@String(ICCV  = {ICCV})

@inproceedings{fomenko2022learning,
    author = {Vladimir Fomenko and Ismail Elezi and Deva Ramanan and Laura Leal-Taix{'e} and Aljo\v{s}a O\v{s}ep},
    title = {Learning to Discover and Detect Objects},
    booktitle={Advances in Neural Information Processing Systems},
    year = {2022}
}

@inproceedings{vaze2022generalized,
  title={Generalized category discovery},
  author={Vaze, Sagar and Han, Kai and Vedaldi, Andrea and Zisserman, Andrew},
  booktitle={Proceedings of the IEEE/CVF Conference on Computer Vision and Pattern Recognition},
  pages={7492--7501},
  year={2022}
}

@inproceedings{rizve2022openldn,
  title={Openldn: Learning to discover novel classes for open-world semi-supervised learning},
  author={Rizve, Mamshad Nayeem and Kardan, Navid and Khan, Salman and Shahbaz Khan, Fahad and Shah, Mubarak},
  booktitle={European Conference on Computer Vision},
  pages={382--401},
  year={2022},
  organization={Springer}
}

@inproceedings{zhang2023promptcal,
  title={Promptcal: Contrastive affinity learning via auxiliary prompts for generalized novel category discovery},
  author={Zhang, Sheng and Khan, Salman and Shen, Zhiqiang and Naseer, Muzammal and Chen, Guangyi and Khan, Fahad Shahbaz},
  booktitle={Proceedings of the IEEE/CVF Conference on Computer Vision and Pattern Recognition},
  pages={3479--3488},
  year={2023}
}

@inproceedings{clip,
  title={Learning Transferable Visual Models From Natural Language Supervision},
  author={Radford, Alec and Kim, Jong Wook and Hallacy, Chris and Ramesh, Aditya and Goh, Gabriel and Agarwal, Sandhini and Sastry, Girish and Askell, Amanda and Mishkin, Pamela and Clark, Jack and Krueger, Gretchen and Sutskever, Ilya},
  booktitle={Proceedings of the 38th International Conference on Machine Learning},
  pages={8748--8763},
  year={2021},
  organization={PMLR}
}

@inproceedings{groundingdino,
  title={Grounding DINO: Marrying DINO with Grounded Pre-Training for Open-Set Object Detection},
  author={Liu, Shilong and Zeng, Zhaoyang and Ren, Tianhe and Li, Feng and Zhang, Hao and Yang, Jie and Li, Chunyuan and Yang, Jianwei and Su, Hang and Zhu, Jun and Zhang, Lei},
  booktitle={European Conference on Computer Vision},
  pages={38--55},
  year={2024},
  organization={Springer}
}

@inproceedings{sam,
  title={Segment Anything},
  author={Kirillov, Alexander and Mintun, Eric and Ravi, Nikhila and Mao, Hanzi and Rolland, Chloe and Gustafson, Laura and Xiao, Tete and Whitehead, Spencer and Berg, Alexander C. and Lo, Wan-Yen and Doll{\'a}r, Piotr and Girshick, Ross},
  booktitle={Proceedings of the IEEE/CVF International Conference on Computer Vision},
  pages={4015--4026},
  year={2023}
}

@misc{dino,
      title={DINO: DETR with Improved DeNoising Anchor Boxes for End-to-End Object Detection}, 
      author={Hao Zhang and Feng Li and Shilong Liu and Lei Zhang and Hang Su and Jun Zhu and Lionel M. Ni and Heung-Yeung Shum},
      year={2022},
      eprint={2203.03605},
      archivePrefix={arXiv},
      primaryClass={cs.CV}
}

@inproceedings{detr,
  title={End-to-End Object Detection with Transformers},
  author={Carion, Nicolas and Massa, Francisco and Synnaeve, Gabriel and Usunier, Nicolas and Kirillov, Alexander and Zagoruyko, Sergey},
  booktitle={European Conference on Computer Vision},
  pages={213--229},
  year={2020},
  organization={Springer}
}

@inproceedings{liu2021swin,
  title={Swin transformer: Hierarchical vision transformer using shifted windows},
  author={Liu, Ze and Lin, Yutong and Cao, Yue and Hu, Han and Wei, Yixuan and Zhang, Zheng and Lin, Stephen and Guo, Baining},
  booktitle={2021 IEEE/CVF international conference on computer vision (ICCV)},
  pages={9992--10002},
  year={2021},
  organization={Ieee}
}

@INPROCEEDINGS{driving-autono,
  author={Chen, Chenyi and Seff, Ari and Kornhauser, Alain and Xiao, Jianxiong},
  booktitle={2015 IEEE International Conference on Computer Vision (ICCV)}, 
  title={DeepDriving: Learning Affordance for Direct Perception in Autonomous Driving}, 
  year={2015},
  volume={},
  number={},
  pages={2722-2730},
  doi={10.1109/ICCV.2015.312}}

@article{wang2023multi,
  title={Multi-modal 3d object detection in autonomous driving: a survey},
  author={Wang, Yingjie and Mao, Qiuyu and Zhu, Hanqi and Deng, Jiajun and Zhang, Yu and Ji, Jianmin and Li, Houqiang and Zhang, Yanyong},
  journal={International Journal of Computer Vision},
  pages={1--31},
  year={2023},
  publisher={Springer}
}

@ARTICLE{od-survey,
  author={Zou, Zhengxia and Chen, Keyan and Shi, Zhenwei and Guo, Yuhong and Ye, Jieping},
  journal={Proceedings of the IEEE}, 
  title={Object Detection in 20 Years: A Survey}, 
  year={2023},
  volume={111},
  number={3},
  pages={257-276},
  doi={10.1109/JPROC.2023.3238524}}

@misc{gupta2019lvis,
      title={LVIS: A Dataset for Large Vocabulary Instance Segmentation}, 
      author={Agrim Gupta and Piotr Dollár and Ross Girshick},
      year={2019},
      eprint={1908.03195},
      archivePrefix={arXiv},
      primaryClass={cs.CV}
}

@misc{wu2023cora,
      title={CORA: Adapting CLIP for Open-Vocabulary Detection with Region Prompting and Anchor Pre-Matching}, 
      author={Xiaoshi Wu and Feng Zhu and Rui Zhao and Hongsheng Li},
      year={2023},
      eprint={2303.13076},
      archivePrefix={arXiv},
      primaryClass={cs.CV}
}

@misc{wu2023aligning,
      title={Aligning Bag of Regions for Open-Vocabulary Object Detection}, 
      author={Size Wu and Wenwei Zhang and Sheng Jin and Wentao Liu and Chen Change Loy},
      year={2023},
      eprint={2302.13996},
      archivePrefix={arXiv},
      primaryClass={cs.CV}
}

@inproceedings{li2021grounded,
      title={Grounded Language-Image Pre-training},
      author={Liunian Harold Li* and Pengchuan Zhang* and Haotian Zhang* and Jianwei Yang and Chunyuan Li and Yiwu Zhong and Lijuan Wang and Lu Yuan and Lei Zhang and Jenq-Neng Hwang and Kai-Wei Chang and Jianfeng Gao},
      year={2022},
      booktitle={CVPR},
}

@misc{du2022vos,
      title={VOS: Learning What You Don't Know by Virtual Outlier Synthesis}, 
      author={Xuefeng Du and Zhaoning Wang and Mu Cai and Yixuan Li},
      year={2022},
      eprint={2202.01197},
      archivePrefix={arXiv},
      primaryClass={cs.LG}
}

@misc{gu2022openvocabulary,
      title={Open-vocabulary Object Detection via Vision and Language Knowledge Distillation}, 
      author={Xiuye Gu and Tsung-Yi Lin and Weicheng Kuo and Yin Cui},
      year={2022},
      eprint={2104.13921},
      archivePrefix={arXiv},
      primaryClass={cs.CV}
}

@incollection{ov-detr,
	doi = {10.1007/978-3-031-20077-9_7},
  
	url = {https://doi.org/10.1007%2F978-3-031-20077-9_7},
  
	year = 2022,
	publisher = {Springer Nature Switzerland},
  
	pages = {106--122},
  
	author = {Yuhang Zang and Wei Li and Kaiyang Zhou and Chen Huang and Chen Change Loy},
  
	title = {Open-Vocabulary {DETR} with~Conditional Matching},
  
	booktitle = {Lecture Notes in Computer Science}
}

@misc{ovr-cnn,
      title={Open-Vocabulary Object Detection Using Captions}, 
      author={Alireza Zareian and Kevin Dela Rosa and Derek Hao Hu and Shih-Fu Chang},
      year={2021},
      eprint={2011.10678},
      archivePrefix={arXiv},
      primaryClass={cs.CV}
}

@misc{detic,
      title={Detecting Twenty-thousand Classes using Image-level Supervision}, 
      author={Xingyi Zhou and Rohit Girdhar and Armand Joulin and Philipp Krähenbühl and Ishan Misra},
      year={2022},
      eprint={2201.02605},
      archivePrefix={arXiv},
      primaryClass={cs.CV}
}

@inproceedings{yolo,
  title={You only look once: Unified, real-time object detection},
  author={Redmon, J},
  booktitle={Proceedings of the IEEE conference on computer vision and pattern recognition},
  year={2016}
}

@inproceedings{coco,
  title={Microsoft coco: Common objects in context},
  author={Lin, Tsung-Yi and Maire, Michael and Belongie, Serge and Hays, James and Perona, Pietro and Ramanan, Deva and Doll{\'a}r, Piotr and Zitnick, C Lawrence},
  booktitle={Computer Vision--ECCV 2014: 13th European Conference, Zurich, Switzerland, September 6-12, 2014, Proceedings, Part V 13},
  pages={740--755},
  year={2014},
  organization={Springer}
}

@inproceedings{li2020overcoming,
  title={Overcoming classifier imbalance for long-tail object detection with balanced group softmax},
  author={Li, Yu and Wang, Tao and Kang, Bingyi and Tang, Sheng and Wang, Chunfeng and Li, Jintao and Feng, Jiashi},
  booktitle={Proceedings of the IEEE/CVF conference on computer vision and pattern recognition},
  pages={10991--11000},
  year={2020}
}

@article{zhao2023survey,
  title={A survey of large language models},
  author={Zhao, Wayne Xin and Zhou, Kun and Li, Junyi and Tang, Tianyi and Wang, Xiaolei and Hou, Yupeng and Min, Yingqian and Zhang, Beichen and Zhang, Junjie and Dong, Zican and others},
  journal={arXiv preprint arXiv:2303.18223},
  year={2023}
}

@inproceedings{yoloworld,
  title={Yolo-world: Real-time open-vocabulary object detection},
  author={Cheng, Tianheng and Song, Lin and Ge, Yixiao and Liu, Wenyu and Wang, Xinggang and Shan, Ying},
  booktitle={Proceedings of the IEEE/CVF Conference on Computer Vision and Pattern Recognition},
  pages={16901--16911},
  year={2024}
}

@inproceedings{rahman2018zero,
  title={Zero-shot object detection: Learning to simultaneously recognize and localize novel concepts},
  author={Rahman, Shafin and Khan, Salman and Porikli, Fatih},
  booktitle={Asian Conference on Computer Vision},
  pages={547--563},
  year={2018},
  organization={Springer}
}

@article{zhao2019object,
  title={Object detection with deep learning: A review},
  author={Zhao, Zhong-Qiu and Zheng, Peng and Xu, Shou-tao and Wu, Xindong},
  journal={IEEE transactions on neural networks and learning systems},
  volume={30},
  number={11},
  pages={3212--3232},
  year={2019},
  publisher={IEEE}
}

@inproceedings{trex2,
  title={T-rex2: Towards generic object detection via text-visual prompt synergy},
  author={Jiang, Qing and Li, Feng and Zeng, Zhaoyang and Ren, Tianhe and Liu, Shilong and Zhang, Lei},
  booktitle={European Conference on Computer Vision},
  pages={38--57},
  year={2025},
  organization={Springer}
}

@article{Ciaglia2022Roboflow1A,
  title={Roboflow 100: A Rich, Multi-Domain Object Detection Benchmark},
  author={Floriana Ciaglia and Francesco Saverio Zuppichini and Paul Guerrie and Mark McQuade and Jacob Solawetz},
  journal={ArXiv},
  year={2022},
  volume={abs/2211.13523},
  url={https://api.semanticscholar.org/CorpusID:254017762}
}

@inproceedings{yao2024detclipv3,
  title={Detclipv3: Towards versatile generative open-vocabulary object detection},
  author={Yao, Lewei and Pi, Renjie and Han, Jianhua and Liang, Xiaodan and Xu, Hang and Zhang, Wei and Li, Zhenguo and Xu, Dan},
  booktitle={Proceedings of the IEEE/CVF Conference on Computer Vision and Pattern Recognition},
  pages={27391--27401},
  year={2024}
}

@article{bharadwaj2023enhancing,
  title={Enhancing Novel Object Detection via Cooperative Foundational Models},
  author={Bharadwaj, Rohit and Naseer, Muzammal and Khan, Salman and Khan, Fahad Shahbaz},
  journal={arXiv preprint arXiv:2311.12068},
  year={2023}
}

@inproceedings{zhao2024taming,
  title={Taming self-training for open-vocabulary object detection},
  author={Zhao, Shiyu and Schulter, Samuel and Zhao, Long and Zhang, Zhixing and Suh, Yumin and Chandraker, Manmohan and Metaxas, Dimitris N and others},
  booktitle={Proceedings of the IEEE/CVF Conference on Computer Vision and Pattern Recognition},
  pages={13938--13947},
  year={2024}
}

@inproceedings{cutler,
  title={CutLER: Cut and Learn for Unsupervised Object Detection and Instance Segmentation},
  author={Wang, Xudong and Girdhar, Rohit and Yu, Stella X. and Misra, Ishan},
  booktitle={Proceedings of the IEEE/CVF Conference on Computer Vision and Pattern Recognition},
  pages={3124--3134},
  year={2023}
}

@inproceedings{freesolo,
  title={FreeSOLO: Learning to Segment Objects without Annotations},
  author={Wang, Xinlong and Yu, Zhiding and De Mello, Shalini and Kautz, Jan and Anandkumar, Anima and Shen, Chunhua and Alvarez, Jose M.},
  booktitle={Proceedings of the IEEE/CVF Conference on Computer Vision and Pattern Recognition},
  pages={14176--14186},
  year={2022}
}

@inproceedings{regionclip,
  title={RegionCLIP: Region-based Language-Image Pretraining},
  author={Zhong, Yiwu and Yang, Jianwei and Zhang, Pengchuan and Li, Chunyuan and Codella, Noel and Li, Liunian Harold and Zhou, Luowei and Dai, Xiyang and Yuan, Lu and Li, Yin and Gao, Jianfeng},
  booktitle={Proceedings of the IEEE/CVF Conference on Computer Vision and Pattern Recognition},
  pages={16793--16803},
  year={2022}
}

@inproceedings{owlv2,
  title={Scaling Open-Vocabulary Object Detection},
  author={Minderer, Matthias and Gritsenko, Alexey and Houlsby, Neil},
  booktitle={Advances in Neural Information Processing Systems},
  volume={36},
  year={2023}
}

@article{groundedsam,
  title={Grounded SAM: Assembling Open-World Models for Diverse Visual Tasks},
  author={Ren, Tianhe and Liu, Shilong and Zeng, Ailing and Lin, Jing and Li, Kunchang and Cao, He and Chen, Jiayu and Huang, Xinyu and Chen, Yukang and Yan, Feng and Zeng, Zhaoyang and Zhang, Hao and Li, Feng and Yang, Jie and Li, Hongyang and Jiang, Qing and Zhang, Lei},
  journal={arXiv preprint arXiv:2401.14159},
  year={2024}
}

@article{maskdistill,
  title={Discovering Object Masks with Transformers for Unsupervised Semantic Segmentation},
  author={Van Gansbeke, Wouter and Vandenhende, Simon and Van Gool, Luc},
  journal={arXiv preprint arXiv:2206.06363},
  year={2022}
}

@inproceedings{tokencut,
  title={Self-Supervised Transformers for Unsupervised Object Discovery Using Normalized Cut},
  author={Wang, Yangtao and Shen, Xi and Hu, Shell Xu and Yuan, Yuan and Crowley, James L. and Vaufreydaz, Dominique},
  booktitle={Proceedings of the IEEE/CVF Conference on Computer Vision and Pattern Recognition},
  pages={14543--14553},
  year={2022}
}
}

\maketitlesupplementary
\setcounter{page}{1}

\section{Additional Experimental Details}

In our proposed work, we demonstrated the performance of the proposed pipeline on novel object detection under domain shift by compiling a dataset of simulated images using a graphic simulator.

In this experiment, we consider the detection performance of the proposed pipeline when changes are introduced to the environment and sensor. To simulate the change in environment, we use an outdoor environment in the simulator. The sensor changes are introduced by the color and texture gap between real and synthetic images. We refer to this cumulative gap between the real images and simulated images as the domain shift.

To study the performance under domain shift, we need to introduce the domain gap during inference from the proposed pipeline. For that, we train the object detector on the COCO dataset as explained in the main paper for the annotator-free object detection task. Then, we use this pre-trained object detector to perform inference on the simulated dataset.

The simulated dataset was constructed using the Applied Intuition simulator. Here, we used five known and novel classes. The known classes: \texttt{car}, \texttt{airplane}, \texttt{bus}, \texttt{truck}, and \texttt{bicycle}, already exist in the COCO categories. The novel classes: \texttt{tank}, \texttt{tractor}, \texttt{helicopter}, \texttt{scooter}, and \texttt{excavator} do not appear in the training distribution. Below we provide some sample images from the simulated dataset.

\begin{figure}[!ht]
    \centering
    \begin{tikzpicture}

        \node (img1) {\includegraphics[width=0.5\linewidth]{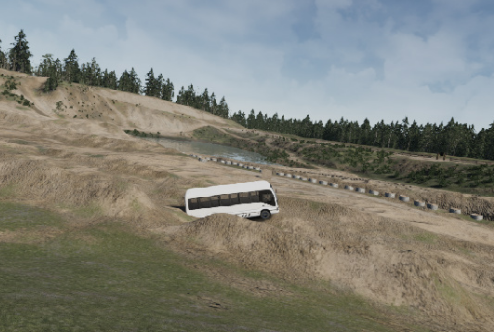}};
        \node[right=of img1, xshift=-1.2cm] (img2) {\includegraphics[width=0.5\linewidth]{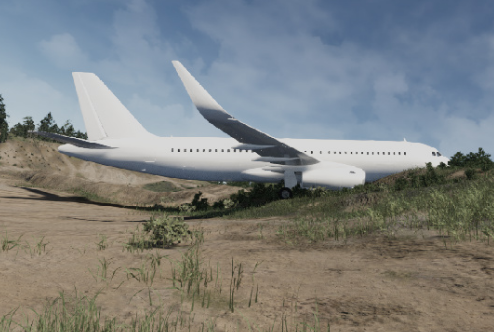}};
        
        \node[below=of img1, anchor=center, yshift=0.9cm] {bus};
        \node[below=of img2, anchor=center, yshift=0.9cm] {airplane};

        \node[below=of img1, yshift=0.65cm] (img3) {\includegraphics[width=0.5\linewidth]{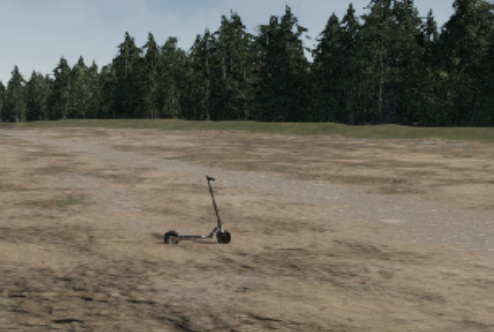}};
        \node[right=of img3, xshift=-1.2cm] (img4) {\includegraphics[width=0.5\linewidth]{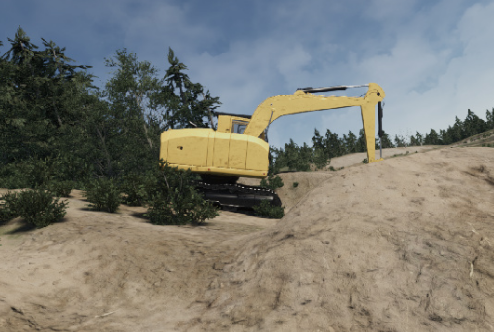}};

        \node[below=of img3, anchor=center, yshift=0.9cm] {scooter};
        \node[below=of img4, anchor=center, yshift=0.9cm] {excavator};

    \end{tikzpicture}

    \caption{Sample images from the simulated dataset.}
\end{figure}

Quantitative comparisons in \cref{tab: domain shift} report results separately for known and novel categories, similar to open-vocabulary evaluations. Our pipeline achieves a clear advantage on novel classes while maintaining strong performance on known categories, driven by the LLM-based classification branch, which supplies a broader and more adaptive vocabulary than conventional open-vocabulary or open-set detectors. Note that relative rankings among baselines on this synthetic benchmark can differ from the main paper's COCO/LVIS/ODinW/Roboflow results: for instance, GLIP-L outperforms T-Rex2 on novel classes here despite the opposite trend there, because this dataset's scene distribution and novel-category composition (e.g., tanks, tractors) differ substantially from those benchmarks.

\begin{table}[!ht]
\centering
\caption{Comparison for novel object detection under domain shift with other state-of-the-art methods. Performance is evaluated on novel, known, and overall classes.}
\label{tab: domain shift}
\begin{tabular}{@{}lcccc@{}}
\toprule
\textbf{Method} & \textbf{Backbone} & \textbf{Novel} & \textbf{Known} & \textbf{Overall} \\
\midrule
GLIP-L \cite{li2021grounded}   & Swin-L  & 35.1 & 12.6	& 24.9 \\
T-Rex2 \cite{trex2}            & Swin-L  & 31.9	& 21.2	& 27.1 \\
GDINO-L \cite{groundingdino}   & Swin-L  & 38.4	& 26.5	& 33.0 \\
Ours                            & YOLO    & \textbf{43.7}	& \textbf{32.8}	& \textbf{38.8} \\
\bottomrule
\end{tabular}
\end{table}

\section{Annotation-Free Detection Pipeline: Clarifications}

\subsection{Pipeline Design and Novelty}

Unlike DetCLIPv3, RegionCLIP, VLDet, or T-Rex2, which all depend on labeled datasets or pre-trained supervised components, our pipeline is target-dataset annotation-free. It is built upon:
\begin{itemize}
    \item \textbf{SAM} for class-agnostic proposal generation,
    \item \textbf{CLIP and LVLMs} for label assignment,
    \item \textbf{Self-refinement} via re-prompting with trained detectors.
\end{itemize}

This design enables detector training without any target-dataset annotations, using only unannotated images. Our modular approach integrates pseudo-labeling, open-world reasoning, and label refinement into a single end-to-end framework.

\subsection{Handling Noisy SAM Proposals}

SAM's automatic mode may produce over- or under-segmented regions. We address this via:
\begin{enumerate}
    \item \textbf{CLIP-based filtering} (threshold = 0.5),
    \item \textbf{NMS filtering} (IoU threshold = 0.6),
    \item \textbf{Self-refinement} with learned detector.
\end{enumerate}

\begin{figure}[h]
    \centering
    \begin{subfigure}[b]{0.14\textwidth}
        \includegraphics[width=\textwidth]{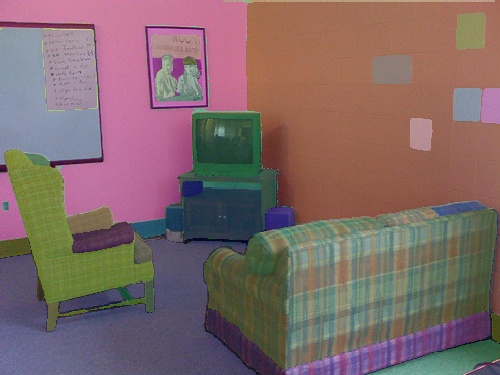}
        \caption{SAM Masks}
    \end{subfigure}
    \hfill
    \begin{subfigure}[b]{0.14\textwidth}
        \includegraphics[width=\textwidth]{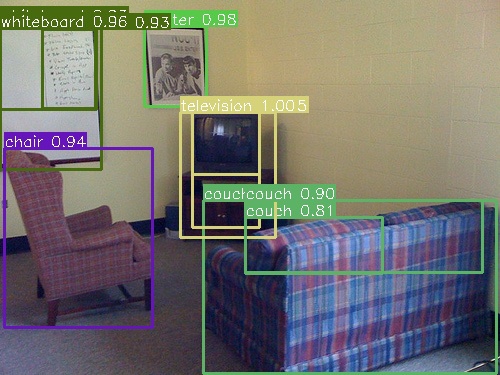}
        \caption{CLIP Filter}
    \end{subfigure}
    \hfill
    \begin{subfigure}[b]{0.14\textwidth}
        \includegraphics[width=\textwidth]{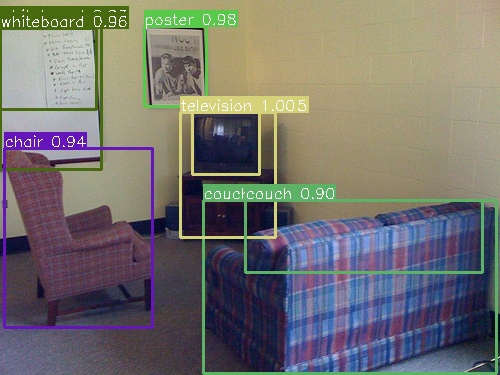}
        \caption{NMS Filter}
    \end{subfigure}
    \caption{Noise reduction steps for SAM proposals.}
\end{figure}

\section{Efficiency and Foundational Model Usage}

Foundational models (CLIP, GPT-4o, CogVLM) are only used during offline proposal labeling. The final deployed detector is a real-time YOLO-based model.

\begin{itemize}
    \item \textbf{CLIP filtering} occurs early to reduce load.
    \item \textbf{GPT-4o} is used only once to generate the label dictionary.
    \item \textbf{LVLMs} are used in refinement (Step 9), limited to fewer high-confidence regions.
\end{itemize}

Average LVLM runtime is 1--2s per image, occurring entirely offline.

\section{Rare Class Generalization}

We evaluated the model on the LVIS rare class subset. The results were:
\begin{itemize}
    \item \textbf{AP\textsubscript{all}} = 47.4
    \item \textbf{AP\textsubscript{r}} = 46.1
\end{itemize}

This small drop suggests better robustness to class imbalance compared to models like T-Rex2.

\section{Scalability of Label Dictionary}

While the label dictionary theoretically incurs \(O(n^2)\) complexity due to transformer attention, we use:
\begin{itemize}
    \item \textbf{Hashmap-based caching} of label outputs,
    \item \textbf{Amortized \(O(1)\)} lookups for repeated scenes.
\end{itemize}

This allows the system to scale efficiently even with large vocabularies.

\section{Parameter Summary}

\begin{itemize}
    \item CLIP Score Threshold (Step 6): 0.5
    \item NMS IoU Threshold: 0.6
    \item LVLMs: Used for semantic refinement, no confidence scores
    \item Detector: YOLO World with default configs
\end{itemize}

\clearpage

\end{document}